\documentclass[conference]{IEEEtran}
\usepackage{times}

\usepackage[numbers]{natbib}
\usepackage{multicol}
\usepackage[bookmarks=true]{hyperref}
\usepackage{graphicx}
\usepackage{algorithm}
\usepackage{algpseudocode}
\usepackage{amssymb}
\usepackage{amsmath}
\usepackage{xcolor}
\usepackage{subcaption}

\newcommand{\algoname}{SAEGT}

\begin{document}

\title{Safe Active Navigation and Exploration for\\ Planetary Environments Using Proprioceptive Measurements}


\author{
  \authorblockN{Matthew Jiang\authorrefmark{1}\authorrefmark{3}, Shipeng Liu\authorrefmark{1}\authorrefmark{3}, Feifei Qian\authorrefmark{1}\authorrefmark{2}}
  \authorblockA{\authorrefmark{1}University of Southern California, Los Angeles, California 90089, USA\\
  \authorrefmark{3}Equal contribution\\
  \authorrefmark{2}Corresponding Author\\
  Email: \{jiangmy, shipengl, feifeiqi\}@usc.edu}
}

\maketitle

\begin{abstract}
Legged robots can sense terrain through force interactions during locomotion, offering more reliable traversability estimates than remote sensing and serving as scouts for guiding wheeled rovers in challenging environments. However, even legged scouts face challenges when traversing highly deformable or unstable terrain. We present Safe Active Exploration for Granular Terrain (\algoname), a navigation framework that enables legged robots to safely explore unknown granular environments using proprioceptive sensing, particularly where visual input fails to capture terrain deformability. \algoname{} estimates the safe region and frontier region online from leg-terrain interactions using Gaussian Process regression for traversability assessment, with a reactive controller for real-time safe exploration and navigation. \algoname{} demonstrated its ability to safely explore and navigate toward a specified goal using only proprioceptively estimated traversability in simulation.
\end{abstract}

\IEEEpeerreviewmaketitle

\section{Introduction}
\label{sec:Introduction}

Planetary exploration often relies on remote or visual sensing to plan safe paths, as in the Mars 2020 rover's use of HazCams and NavCams \cite{maki2020mars2020}. However, failures like the Spirit rover's entrapment \cite{callas2015spirit} reveal that vision systems cannot capture subsurface properties of deformable terrains. In contrast, legged robots use proprioceptive feedback \cite{liu2025adaptive, 2024LPICo3040.1806F, 2023LPICo2806.1733B, 2024LPICo3040.1798B} during each step to directly assess terrain properties, offering sensing capabilities that are unaffected by the limitations of vision-based systems. This makes them well-suited for scouting roles in planetary exploration. However, despite their proven mobility in unstructured~\cite{hwangbo2019learning, lee2024learning} and soft granular terrains~\cite{qian2013walking, liu2023adaptation, li2009sensitive}, legged robots still face locomotion challenges on the soft and cohesive surfaces \cite{morphology1979dunes} of planetary environments. This highlights the need for a safe and active navigation framework to support their role as scouts with only point-sensor measurements in planetary exploration.

To enable legged robots to navigate and actively scout based on proprioceptive measurements, this work introduces a safe navigation framework composed of two main components. We first employ a Gaussian Process using point-wise proprioceptive measurements to create a safe region map, where we know with a tunable level of confidence that the robot can walk safely, and a frontier map indicating where the robot can expand the safe region by exploring further. Second, we combine these regions with a reactive, geometry-based local controller, \algoname, enabling legged robots to actively map terrain mechanical properties, update safe and frontier regions in real time, and safely explore while supporting other navigation objectives.

\section{Related work}
Robot exploration and navigation frameworks generally can be divided into two categories: geometric-based~\cite{leininger2024gaussianprocessbasedtraversabilityanalysis} and learning-based approaches~\cite{weerakoon2022terp}. In this work, we are interested in the first one. These approaches often require the robot to construct a map from exteroceptive sensors (cameras, LiDAR)~\cite{uttsha2024gaussianprocessdistancefields} and analyze the traversability. Gaussian Process-based methods have emerged as powerful tools to estimate a continuous map~\cite{leininger2024gaussianprocessbasedtraversabilityanalysis}, and traversability is then often analyzed based on slope, roughness, and step height~\cite{wermelinger2016navigation, wang2023towards}. However, on deformable terrains, the robot must rely on proprioceptive measurements to assess terrain properties, which can only be obtained upon contact. This paper presents a Gaussian Process-based method to generate a safe region and a frontier region, enabling the robot to actively map and explore using proprioceptive feedback.

After generating the traversability map, a planner is needed to compute the robot's path. If the goal is to efficiently map the environment, the planner seeks to maximize information entropy~\cite{ghaffari2018gaussian, ali2023autonomous, jang2020multi}; if there is a global goal, it focuses on ensuring safe traversal to the goal location. Two main approaches are often used: (1) plan a trajectory using methods such as RRT~\cite{lavalle1998rapidly}, RRT*~\cite{karaman2011sampling}, Dijkstra~\cite{dijkstra2022note}, or A*~\cite{hart1968formal}, followed by a controller to track the trajectory; or (2) directly apply potential field-based methods to convert the occupancy map into a vector field to control the robot~\cite{arslan2016exactrobotnavigation, arslan2016sensorbasedreactive, Vasilopoulos_RAL_2020, vasilopoulos2021reactivenavigationpartiallyfamiliar}. Our method integrates the identified safe region and frontier region with potential-field-based methods, directly generating velocity control commands for the robot to either navigate to a global goal or explore the environment.

\section{Proprioceptive traversability mapping for granular terrains}
In this section, we present a granular-terrain traversability mapping framework composed of two tightly integrated components: (i) Gaussian Process-based traversability mapping from per-step proprioceptive sensing, which incrementally constructs a global traversability map $\mathcal{T}_t$; and (ii) region detection and expansion using a Bayesian safety optimization algorithm, which iteratively expands the safe region $S_t$ by leveraging both $\mathcal{T}_t$ and the associated uncertainty estimates. This process simultaneously identifies the associated frontier region $G_t$ for targeted exploration. The frontier points, $\mathbf{g}_t\in G_t$, are potential targeted exploration goals whose traversal can enlarge the robot's safe region, $S_t$.

\subsection{Proprioceptive Traversability Mapping}
\label{sec:ProprioceptiveMapping}
The robot operates as a point at position $\mathbf{x}_t \in \mathbb{R}^2$ within workspace $\mathcal{W}$, discretized into grid $D$ with resolution $\delta$. The proprioceptive system infers terrain traversability $y_t$ through leg-terrain interactions~\cite{liu2025adaptive, marteau2023initial}, combined with terrain dynamics models~\cite{li2009sensitive} to predict entrapment risk~\cite{nevatia2013improved}. 

The traversability function $f(\mathbf{x})$ represents a scalar score derived from the terrain dynamics model's assessment of mechanical properties. Higher values of $f(\mathbf{x})$ correspond to terrain with lower predicted risk of entrapment or mobility failure. We assume traversability measurements are available as noisy observations at each location $\mathbf{x}_t$:
\begin{equation}
y_t = f(\mathbf{x}_t) + \varepsilon_t
\end{equation}
where $f(\mathbf{x}_t) \in \mathbb{R}$ represents the true terrain traversability at location $\mathbf{x}_t$, and $\varepsilon_t \sim \mathcal{N}(0, \sigma_\text{noise}^2)$ accounts for measurement noise and processing uncertainty in the proprioceptive sensing and terrain dynamics modeling pipeline.

The terrain traversability map $\mathcal{T}_t$ is maintained as a Gaussian Process posterior over measurement history $\mathcal{D}_t = \{(\mathbf{x}_i, y_i)\}_{i=1}^t$, using the RBF kernel:
\begin{equation}
k(\mathbf{x}, \mathbf{x}') = \sigma_f^2 \exp\left(-\frac{\|\mathbf{x} - \mathbf{x}'\|^2}{2\ell^2}\right)
\end{equation}
The GP provides predictive mean $\mu_t(\mathbf{x})$ and variance $\sigma_t^2(\mathbf{x})$ at any location $\mathbf{x}$ of the terrain traversability, yielding both terrain traversability predictions and uncertainty estimates that guide safe exploration.

\subsection{Region Detection}
\label{sec:ConfidenceGuided}

\subsubsection{Traversability Confidence Interval Construction}
The Gaussian Process traversability mapping provides predictive mean $\mu_t(\mathbf{x})$ and uncertainty $\sigma_t(\mathbf{x})$. To determine safe and frontier regions, we construct confidence bounds on the predicted traversability values at each position $\mathbf{x}$.

A confidence parameter $\beta$ controls the exploration-safety trade-off, with lower values corresponding to more conservative safety assessments (e.g., $\beta = 1$ corresponds to approximately 68\% confidence \cite{neyman1937outline}):
\begin{equation}
c_t(\mathbf{x}) = \left[\mu_{t-1}(\mathbf{x}) \pm \sqrt{\beta}\, \sigma_{t-1}(\mathbf{x})\right]
\end{equation}

The confidence bounds are iteratively intersected with prior confidence sets to ensure consistency:
\begin{equation}
    C_t(\mathbf{x}) = C_{t-1}(\mathbf{x}) \cap c_t(\mathbf{x})
\end{equation}
From $C_t(\mathbf{x})$, the lower and upper bounds of traversability are defined as:
\begin{equation}
    \ell_t(\mathbf{x}) = \min C_t(\mathbf{x}), \quad u_t(\mathbf{x}) = \max C_t(\mathbf{x})
\end{equation}

\subsubsection{Safe Region Detection}
Following previous work \cite{safeopt}, the lower confidence bounds are used to iteratively expand regions that are safe with high probability for the robot to traverse. 

The safe region expansion relies on the assumption that the true traversability $f(\mathbf{x})$ is Lipschitz continuous with constant $L$, meaning $|f(\mathbf{x}) - f(\mathbf{x}')| \leq L\|\mathbf{x} - \mathbf{x}'\|$ for all points $\mathbf{x}, \mathbf{x}'$ in the workspace \cite{royden2010real}. For granular terrains, this reflects the physical notion that soil mechanical properties change gradually over space due to similar formation processes, loading history, and environmental conditions affecting neighboring regions. The safe region $S_t$ is constructed by expanding from previously safe regions:
\begin{equation}
S_t = \bigcup_{\mathbf{x} \in S_{t-1}} \left\{ \mathbf{x}' \in D \,\middle|\, \ell_t(\mathbf{x}) - L \|\mathbf{x}-\mathbf{x}'\| \geq h \right\}
\end{equation}
where $\ell_t(\mathbf{x})$ is the lower confidence bound of traversability, $h$ is the safety threshold, and $L$ is the Lipschitz constant of the traversability function $f(\mathbf{x})$. 

The robot begins with an initial small safe region $S_0$ based on its starting location and initial measurements. This region is then expanded iteratively as new measurements reduce uncertainty and allow the lower confidence bounds to exceed the safety threshold in previously unknown regions.

\subsubsection{Frontier Region Detection}

To identify regions that could potentially expand the safe region, we utilize the expansion objective/acquisition function \cite{safeopt}:
\begin{equation}
    g_t(\mathbf{x}) = \left| \left\{ \mathbf{x}' \in D \setminus S_t : u_t(\mathbf{x}) - L \|\mathbf{x} - \mathbf{x}'\| \geq h \right\} \right|
\end{equation}
which evaluates how large the existing safe region could expand if point $\mathbf{x}$ is sampled. Using the upper confidence bound $u_t(\mathbf{x})$, the potential safe region expansion is evaluated under the most favorable traversability scenario. To consider the expansion objective, let the frontier region $G_t$ consist of safe points with positive expansion potential:
\begin{equation}
    G_t = \{\mathbf{x}\in S_t \mid g_t(\mathbf{x}) > 0\}
\end{equation}

These frontier points represent candidate sampling locations that could enlarge the verified safe region when visited.

\section{Reactive Navigation For Safe Exploration}
This section outlines the navigation strategy of SAEGT, which first selects a frontier subgoal that balances exploration and progress toward the goal. It then generates a potential field based on the selected subgoal and traversability prediction to guide the robot in safely exploring the environment.

\subsection{Frontier Subgoal Selection}

The robot selects the next sampling point by balancing two objectives:
\begin{itemize}
    \item Expand the safe regions
    \item Make progress toward the mission goal
\end{itemize}

After identifying the safe region $S_t$ and the frontier region $G_t \subseteq S_t$, the robot employs a hybrid strategy to guide exploration in a goal-directed yet uncertainty-aware manner.

\subsubsection{Subgoal Selection Procedure}

The subgoal selection procedure is as follows:
\begin{enumerate}
    \item For each point $\mathbf{x} \in G_t$, compute its distance to the goal $\|\mathbf{x} - \mathbf{x}_g\|$.
    \item Sort all points in $G_t$ in ascending order of distance to the goal.
    \item Consider only the top $n$ nearest frontier points, where $n$ is a configurable parameter.
    \item Among these $n$ frontier points, select the point with the largest confidence interval width $w_t(\mathbf{x}) := u_t(\mathbf{x}) - \ell_t(\mathbf{x})$.
\end{enumerate}

This approach prioritizes frontier points that offer more information, while guiding the robot toward its goal using a distance heuristic. It mitigates the risk of wasting samples on irrelevant directions and prevents purely greedy selection based on proximity alone, which might lead into regions of high uncertainty without expanding the safe region.

\begin{algorithm}[H]
\caption{Safe Active Exploration for Granular Terrain (SAEGT)}
\begin{algorithmic}[1]
\Require Robot workspace $D$, GP prior $(\mu_0, k, \sigma_0)$, Lipschitz constant $L$, initial safe region $S_0$, safety threshold $h$, goal location $\mathbf{x}_g$, number of frontier points to consider $n$
\State $C_0(\mathbf{x}) \gets [h, \infty), \, \forall \mathbf{x} \in S_0$
\State $C_0(\mathbf{x}) \gets \mathbb{R}, \, \forall \mathbf{x} \in D \setminus S_0$
\State $c_0(\mathbf{x}) \gets \mathbb{R}, \, \forall \mathbf{x} \in D$
\For{each time step $t = 1, 2, \ldots$}
    \State $C_t(\mathbf{x}) \gets C_{t-1}(\mathbf{x}) \cap c_{t-1}(\mathbf{x})$
    \State $S_t \gets \bigcup_{\mathbf{x} \in S_{t-1}} \left\{ \mathbf{x}' \in D \,\middle|\, \ell_t(\mathbf{x}) - L \|\mathbf{x}-\mathbf{x}'\| \geq h \right\}$
    \State $G_t \gets \left\{ \mathbf{x} \in S_t \,\middle|\, g_t(\mathbf{x}) > 0 \right\}$
    \State Sort $G_t$ by $\|\mathbf{x} - \mathbf{x}_g\|$ in ascending order
    \State Let $G_t^{(n)} \subseteq G_t$ be the top $n$ nearest frontier points
    \State $\mathbf{x}_t \gets \arg\max_{\mathbf{x} \in G_t^{(n)}} w_t(\mathbf{x})$
    \State Navigate to $\mathbf{x}_t$ using reactive controller
    \State Observe $y_t = f(\mathbf{x}_t) + \varepsilon_t$
    \State Update GP with $(\mathbf{x}_t, y_t)$
    \State Compute $c_t(\mathbf{x})$ for all $\mathbf{x} \in D$
\EndFor
\end{algorithmic}
\end{algorithm}

This strategy combines uncertainty-driven sampling with spatial prioritization to ensure that each move both refines the terrain model and progresses toward the mission objective.

\subsection{Reactive Navigation}

While the planner identifies target locations, executing safe motion between sampling points requires a robust local navigation strategy. Classical global planners require full replanning after environmental updates, making them unsuitable for dynamic terrain scenarios. Instead, we implement a reactive control framework adapted from \citet{vasilopoulos2021reactivenavigationpartiallyfamiliar}.

\subsubsection{Geometric Obstacle Construction}

To enable reactive navigation, the robot constructs geometric representations of safe and hazardous terrain from its evolving traversability map. This process transforms pointwise traversability predictions into polygonal approximations suitable for geometry-aware motion planning.

\begin{figure}[h]
    \centering
    \begin{subfigure}{0.4\linewidth}
        \includegraphics[width=\textwidth]{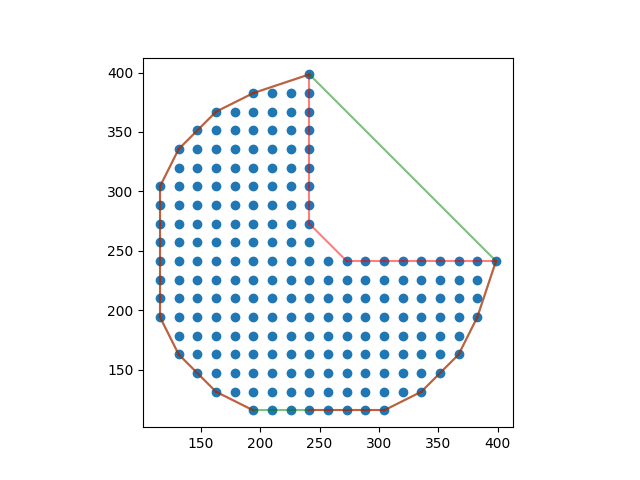}
        \caption{}
    \end{subfigure}
    \hfill
    \begin{subfigure}{0.55\linewidth}
        \includegraphics[width=\textwidth]{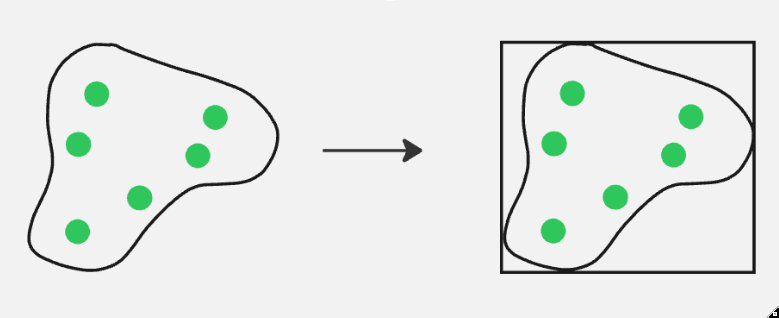}
        \caption{}
    \end{subfigure}
    \caption{Geometric processing pipeline for converting safe terrain points to navigable free space representations. (a) Visualization of a concave hull generated from point clusters. Image adapted from \citet{cubao2024concavehull}. Licensed under the MIT License. (b) Bounding box creation for workspace definition. }
    \label{fig:geometric-processing}
\end{figure} 

The conversion proceeds as follows:
\begin{enumerate}
    \item \textbf{Cluster Formation:} The safe set $S_t$ is partitioned into disjoint spatial clusters using KD-tree \cite{10.1145/361002.361007} nearest-neighbor queries. Each cluster represents a contiguous region of traversable terrain.

    \item \textbf{Concave Hull Construction:} For each cluster, a concave hull is generated using the algorithm from \cite{park2013concavehull}, with parameters controlling shape resolution and computational complexity.

    \item \textbf{Workspace Definition:} A bounding rectangle $\mathcal{W}_e$ encloses all hull regions:
    \begin{equation}
    \mathcal{W}_e := \left\{ \mathbf{x} \in \mathbb{R}^2 \,\middle|\, \underline{H}_t \leq \mathbf{x} \leq \overline{H}_t \right\}
    \end{equation}

    \item \textbf{Obstacle Extraction:} Regions not covered by safe hulls become pseudo-physical obstacles:
    \begin{equation}
    \mathcal{O}_t := \mathcal{W}_e \setminus \bigcup_{i} H_{ti}
    \end{equation}
\end{enumerate}

\subsubsection{Diffeomorphic Reactive Control}

We adopt the reactive navigation framework from \citet{vasilopoulos2021reactivenavigationpartiallyfamiliar}, which constructs a diffeomorphic transformation between the physical workspace and a simplified model space. This enables obstacle avoidance using simple potential fields in the transformed domain. The method guarantees that smooth trajectories in model space correspond to safe, obstacle-free paths in physical space, provided the obstacle set is polygonal. This allows real-time control generation without full replanning as the known obstacle set evolves.

\subsection{System Integration}

The final system integrates global terrain expansion and local reactive control into a unified exploration loop: proprioceptive measurements update the GP-based terrain model, the planner selects new targets from safe regions, and the reactive controller executes real-time movement.

This dual-loop architecture allows the robot to safely explore unknown, deformable environments while efficiently progressing toward its scientific objectives.

\section{Results}
\label{sec:Results}

This section presents simulation-based validation of \algoname\ across terrain scenarios that demonstrate its core capabilities in two scenarios: exploration with a certain goal/priority, and exploratory mapping without preassigned goals.

\subsection{Simulation Setup}

The robot is initialized within a verified safe region with randomly sampled proprioceptive measurements to bootstrap the Gaussian Process model. Test environments are represented as 2D grids where each cell contains terrain traversability values, with the traversability function \( f(\mathbf{x}) \) encoding relative terrain strength and safety threshold \( h \) defining traversability boundaries (e.g., \( h = 1000 \) for demonstration). $\beta = 3$ was used, meaning a $99.7\%$ confidence interval was used.

\begin{figure}[h]
    \centering
    \includegraphics[width=0.7\linewidth]{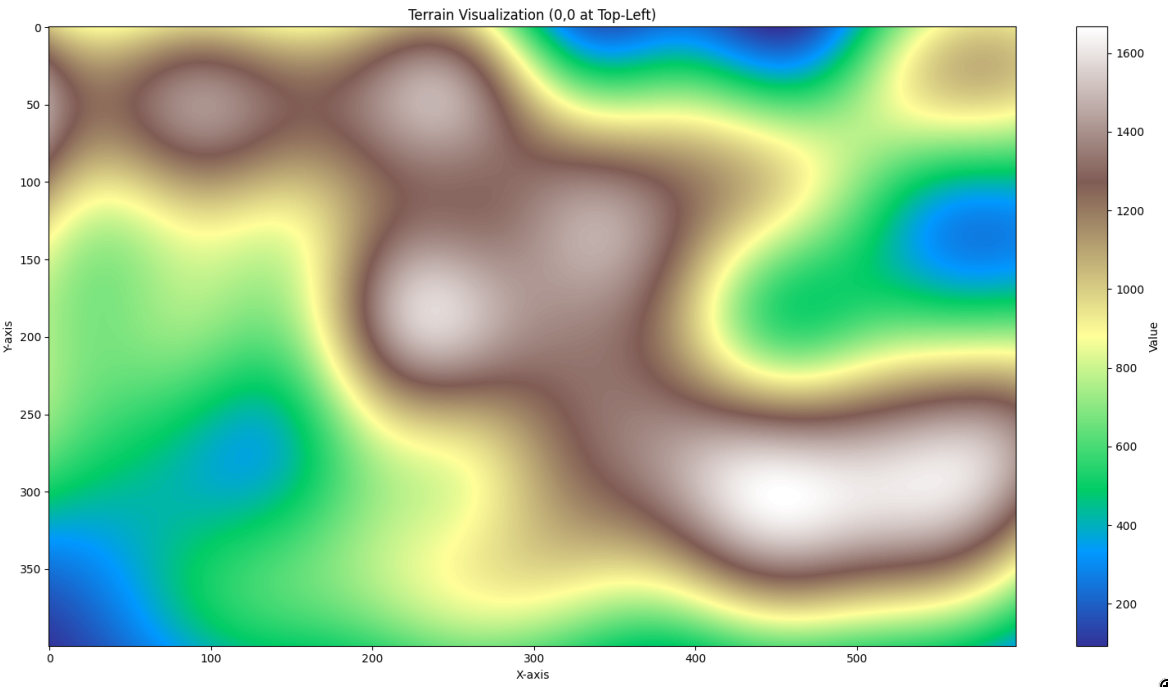}
    \caption{Ground truth traversability values for experiments.}
    \label{fig:terrain}
\end{figure}

\subsection{Experimental Results}

\begin{figure}[h]
    \centering
    \includegraphics[width=0.75\linewidth]{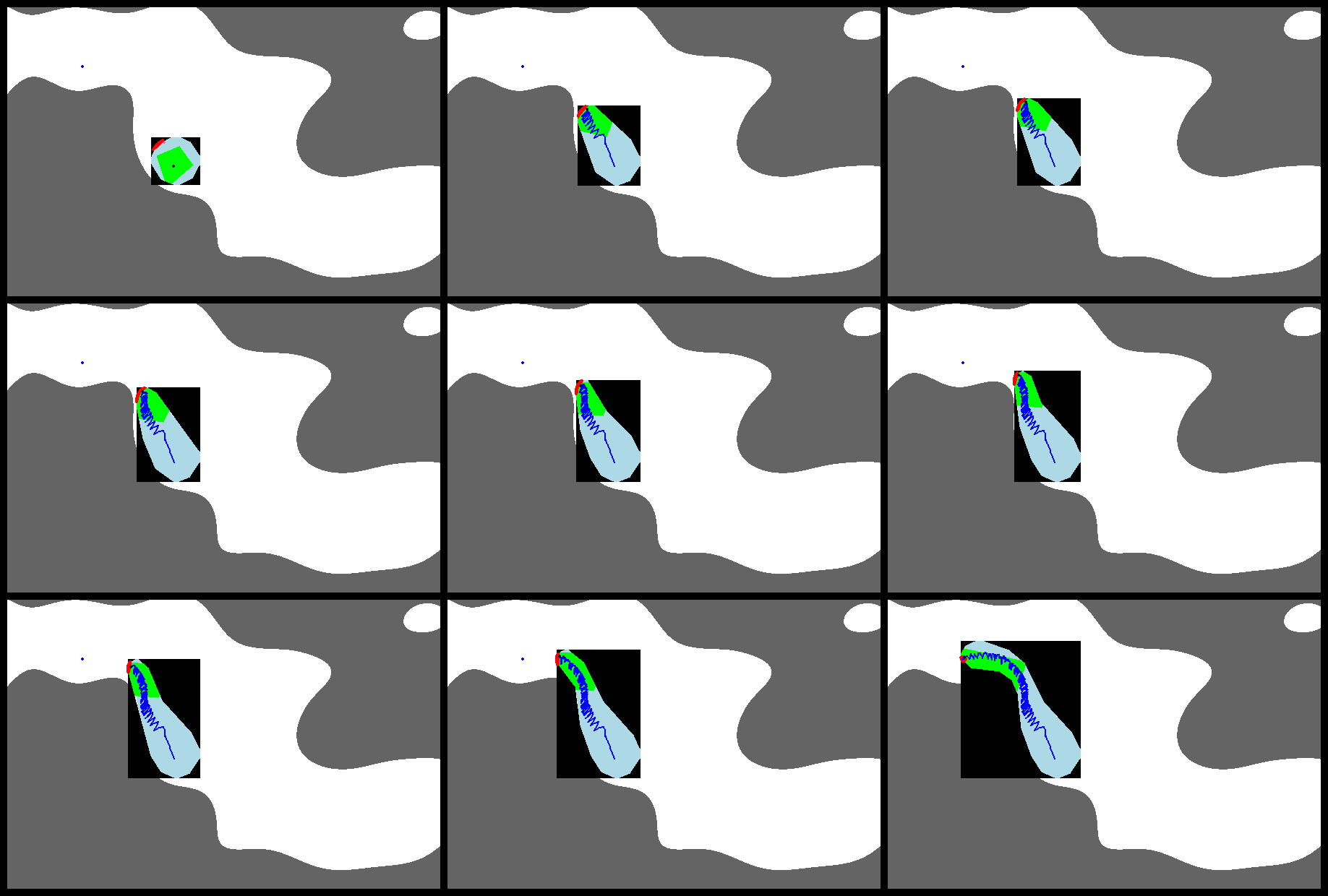}
    \caption{Navigation around low-traversability zone (\( h = 1000 \))}
    \label{fig:navigation}
\end{figure}

\paragraph{Navigation Around Low-Traversability Zones}
The robot successfully navigates around low-traversability regions by iteratively expanding the safe region until reaching goals beyond obstacles. Two scenarios were completed in 210 and 461 iterations, respectively (\autoref{fig:navigation}). The algorithm initially expands rapidly, then explores cautiously around low-traversability peripheries, and finally accelerates once past dangerous areas.

\begin{figure}[h]
    \centering
    \includegraphics[width=0.75\linewidth]{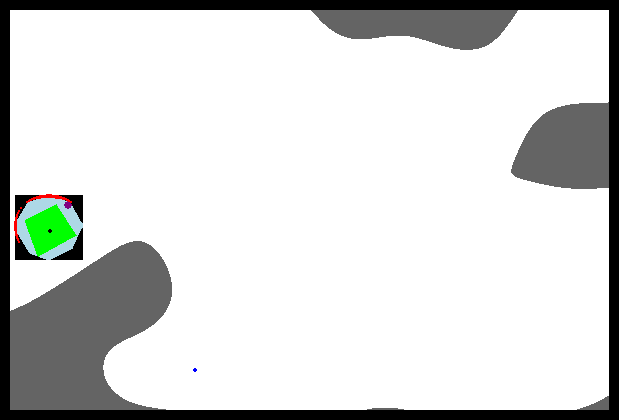}
    \caption[A single frame showing the navigation regions and setup. White = safe terrain (\( f(\mathbf{x}) \geq h \)); Gray = low-traversability terrain (\( f(\mathbf{x}) < h \)); Black = obstacles (\( \mathcal{O}_t \)); Green = local freespace; Light Blue = safe region (\( S_t \)); Red = frontier region (\( G_t \)); Purple = intermediate goal; Dark Blue = final goal.]{A single frame showing the navigation regions and setup.\footnotemark{} White = safe terrain (\( f(\mathbf{x}) \geq h \)); Gray = low-traversability terrain (\( f(\mathbf{x}) < h \)); Black = obstacles (\( \mathcal{O}_t \)); Green = local freespace; Light Blue = safe region (\( S_t \)); Red = frontier region (\( G_t \)); Purple = intermediate goal; Dark Blue = final goal.}
    \label{fig:single}
\end{figure}

\footnotetext{It is important to note that the gray and white areas are shown as ground truth to validate the algorithm, but are not information known to the robot during navigation, while the contents surrounded by the enclosing workspace (outlined by the black obstacles) \textit{are} known and are created by \algoname.}

\begin{figure}[hb]
    \centering
    \includegraphics[width=0.8\linewidth]{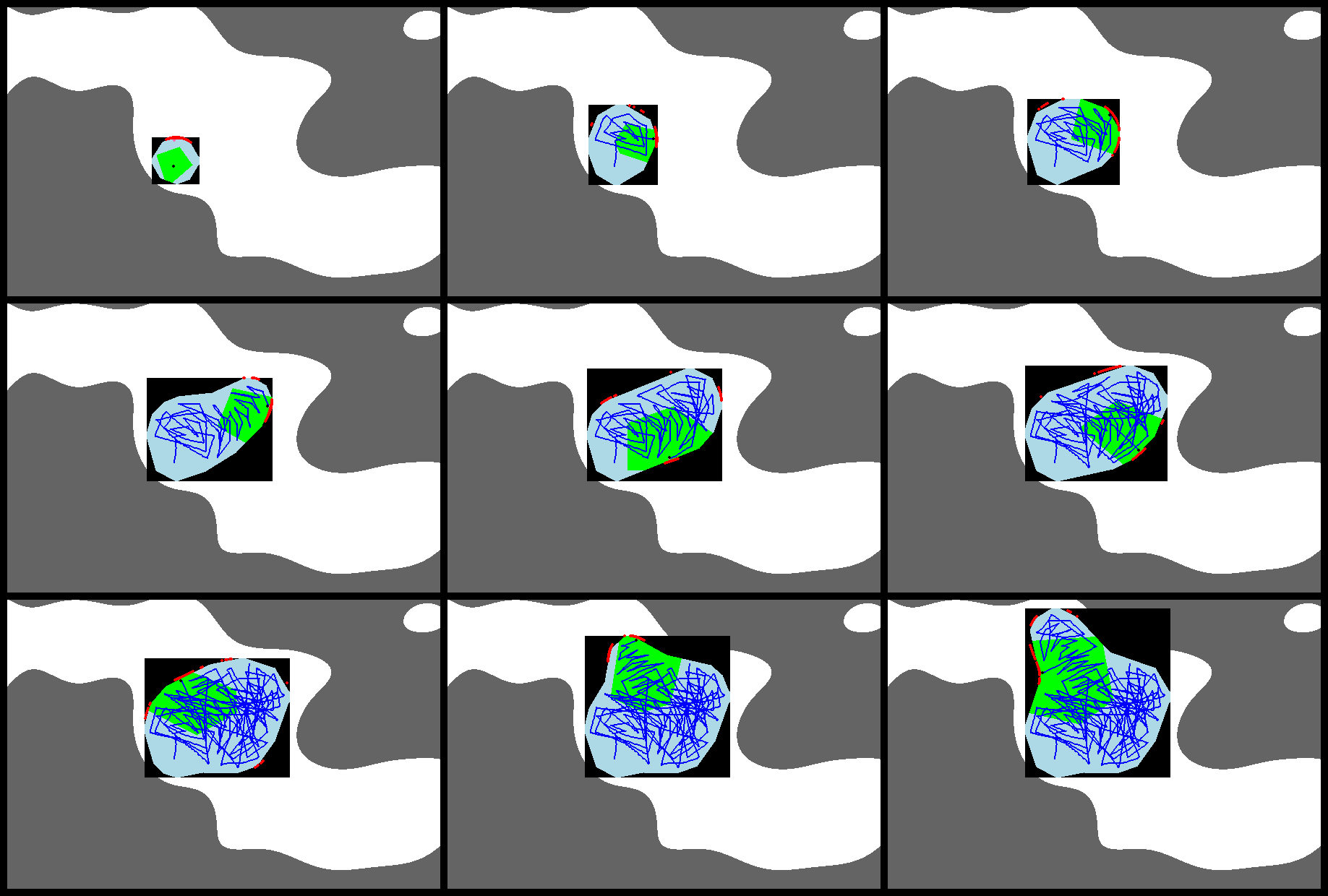}
    \caption{Overall exploration progression over 9 intervals.}
    \label{fig:explorefull}
\end{figure}

\paragraph{Goal-Free Exploration}
Without predefined destinations, the robot selects points maximizing uncertainty reduction, rapidly expanding the known safe area (\autoref{fig:explorefull}). This demonstrates \algoname's capability for general environmental mapping, useful for initial site characterization before goal-directed missions.

The simulation experiments presented validate the intended capabilities of \algoname\ under a variety of terrain conditions and traversability scenarios. The robot successfully performs uncertainty-aware navigation around hazardous terrain, reuses previously acquired safe knowledge to complete tasks more efficiently, and performs autonomous environmental mapping using only local proprioceptive measurements.

\section{Conclusion}
\label{sec:Conclusion}

This paper presents \algoname, a navigation framework for proprioceptive terrain-aware exploration in granular environments where visual sensing is unreliable. Designed to address uncertainty, deformability, and incomplete knowledge in planetary terrains, \algoname\ combines GP-based terrain modeling, confidence-guided safe region expansion, and reactive control to enable legged robots to explore safely using only local physical interaction. 



\clearpage
\bibliographystyle{plainnat}
\bibliography{references}

\clearpage
\appendix
\label{sec:appendix}

This appendix provides detailed explanations for the key steps in the \algoname~algorithm presented in Algorithm 1.

\subsection{Algorithm Step Annotations}

\textbf{Line 5: Update traversability confidence} - This step intersects the current confidence interval with the previous one to maintain consistency across time steps. The intersection ensures that confidence bounds become progressively tighter as more measurements are collected.

\textbf{Line 6: Expand safe region} - The safe region expansion uses the Lipschitz continuity assumption to propagate safety guarantees from previously verified safe points to nearby locations. Points are added to the safe region if their lower confidence bound, adjusted for distance-based uncertainty, exceeds the safety threshold.

\textbf{Line 7: Identify frontier region} - Frontier points are identified as safe locations that have the potential to expand the safe region if visited. These points represent the boundary between known safe terrain and unexplored areas.

\textbf{Line 8: Prioritize by goal distance} - Frontier points are sorted by their Euclidean distance to the goal location to bias exploration toward the mission objective while maintaining safety.

\textbf{Line 10: Select most informative} - Among the nearest frontier candidates, the algorithm selects the point with the largest confidence interval width, maximizing information gain from the next measurement.

\textbf{Line 12: Collect proprioceptive measurement} - The robot obtains a noisy observation of terrain traversability through physical interaction with the terrain at the selected location.

\end{document}